\documentclass[11pt]{article}

\usepackage{arabtex}
\usepackage{utf8}
\setcode{utf8}
\newcommand{\biggestnumber}[1]{\textbf{#1}}
\usepackage{booktabs}
\usepackage{multirow}
\usepackage{amsfonts}
\usepackage[preprint]{acl}
\usepackage{graphicx}
\usepackage{float}
\usepackage{amsmath}
\usepackage{amstext}
\usepackage{enumitem}
\usepackage{listings}
\usepackage{pgf}
\usepackage{tikz}
\usepackage{xcolor}
\usepackage{caption}
\usetikzlibrary{shapes.geometric}

\usepackage{microtype}
\usepackage[most]{tcolorbox}
\usepackage{titling}

\makeatletter
\pgfkeys{/tikz/.cd,
  dogear size/.initial=20pt,
}

\pgfdeclareshape{dogeared}{
  \inheritsavedanchors[from=rectangle]
  \inheritanchorborder[from=rectangle]
  \inheritanchor[from=rectangle]{center}
  \inheritanchor[from=rectangle]{north}
  \inheritanchor[from=rectangle]{south}
  \inheritanchor[from=rectangle]{west}
  \inheritanchor[from=rectangle]{east}

  \savedmacro\dogearsize{%
    \edef\dogearsize{\pgfkeysvalueof{/tikz/dogear size}}%
  }

  \backgroundpath{
    \southwest \pgf@xa=\pgf@x \pgf@ya=\pgf@y
    \northeast \pgf@xb=\pgf@x \pgf@yb=\pgf@y
    \pgf@xc=\pgf@xb \advance\pgf@xc by-\dogearsize
    \pgf@yc=\pgf@yb \advance\pgf@yc by-\dogearsize
    \pgfpathmoveto{\pgfpoint{\pgf@xa}{\pgf@ya}}
    \pgfpathlineto{\pgfpoint{\pgf@xa}{\pgf@yb}}
    \pgfpathlineto{\pgfpoint{\pgf@xc}{\pgf@yb}}
    \pgfpathlineto{\pgfpoint{\pgf@xb}{\pgf@yc}}
    \pgfpathlineto{\pgfpoint{\pgf@xb}{\pgf@ya}}
    \pgfpathclose
  }

  \foregroundpath{
    \southwest \pgf@xa=\pgf@x \pgf@ya=\pgf@y
    \northeast \pgf@xb=\pgf@x \pgf@yb=\pgf@y
    \pgf@xc=\pgf@xb \advance\pgf@xc by-\dogearsize
    \pgf@yc=\pgf@yb \advance\pgf@yc by-\dogearsize
    \pgfpathmoveto{\pgfpoint{\pgf@xc}{\pgf@yb}}
    \pgfpathlineto{\pgfpoint{\pgf@xc}{\pgf@yc}}
    \pgfpathlineto{\pgfpoint{\pgf@xb}{\pgf@yc}}
    \pgfpathclose
  }
}
\makeatother

\tcbset{
  dogear box/.style={
    enhanced,
    colback=white,
    colframe=black,
    width=0.95\textwidth,
    boxrule=0.4pt,
    arc=2pt,
    left=6pt,
    right=6pt,
    top=6pt,
    bottom=6pt,
    overlay={
      \path[fill=gray!10, draw=black]
        (frame.north east)                     % top-right
        -- ++(-10pt,-10pt)                     % toward center
        -- ++(10pt,0pt)                        % right
        -- cycle;
    }
  }
}

\title{\raisebox{-0.1cm}{\includegraphics[height=.86cm]{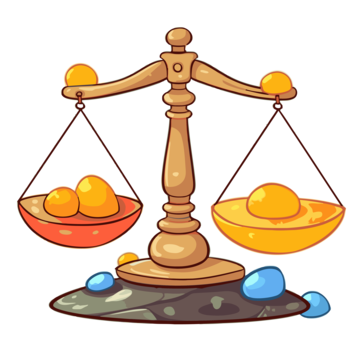}} MELAC: Massive Evaluation of Large  Language Models with Alignment of Culture in Persian Language}

\author{
    \textbf{Farhan Farsi\textsuperscript{1}}\thanks{\; Equal contribution.},
    \textbf{Farnaz Aghababaloo\textsuperscript{2}}\footnotemark[1],
    \textbf{Shahriar Shariati Motlagh\textsuperscript{3}}\footnotemark[1],
    \textbf{Parsa Ghofrani\textsuperscript{1}},
  \\
    \textbf{MohammadAli SadraeiJavaheri\textsuperscript{2}},
    \textbf{Shayan Bali\textsuperscript{4}},
    \textbf{Amirhossein Shabani\textsuperscript{2}},
    \textbf{Farbod Bijary\textsuperscript{1}},
\\
    \textbf{Ghazal Zamaninejad\textsuperscript{2}},
    \textbf{AmirMohammad Salehoof\textsuperscript{2}},
    \textbf{Saeedeh Momtazi\textsuperscript{1}}
\\
    \textsuperscript{1}Amirkabir University of Technology,
    \textsuperscript{2}Part AI Research Center,
    \textsuperscript{3}University of Mazandaran,
\\ 
    \textsuperscript{4}King’s College London 
\\
    \small \{farhan1379, parsa.ghofrani, farbod.bijary, momtazi\}@aut.ac.ir, s.shariati21@umail.umz.ac.ir, shayan.bali@kcl.ac.uk, \\ 
    \small \{farnaz.babalou, mohammad.sadraei, amirhosein.shabani, ghazal.zamaninezhad, amirmohammad.salehoof\}@partdp.ai
}
   
\begin{document}
\maketitle
\begin{abstract}
As large language models (LLMs) become increasingly embedded in our daily lives, evaluating their quality and reliability across diverse contexts has become essential. While comprehensive benchmarks exist for assessing LLM performance in English, there remains a significant gap in evaluation resources for other languages. Moreover, because most LLMs are trained primarily on data rooted in European and American cultures, they often lack familiarity with non-Western cultural contexts. To address this limitation, our study focuses on the Persian language and Iranian culture. We introduce 19 new evaluation datasets specifically designed to assess LLMs on topics such as Iranian law, Persian grammar, Persian idioms, and university entrance exams. Using these datasets, we benchmarked 41 prominent LLMs, aiming to bridge the existing cultural and linguistic evaluation gap in the field.
\end{abstract}

\section{Introduction}
Large Language Models (LLMs) have experienced significant advancements in recent years, including in real-world applications, even those that require in-field expertise, such as software development \cite{jimenez2023swe,sabouri2025trust}, law \cite{sun2024lawluo,cheong2024not},
medical science \cite{goyal2024healai,kim2023chatgpt}, and
religious studies \cite{trepczynski2023religion}.
Researchers attributed this surprising enhancement to emerging capabilities that happen in bigger models during training~\cite{wei2022emergent}.

Nowadays, exploring what LLMs can't do, as opposed to what they can do, has become an interesting topic of study which sheds light on future development~\cite{chen2024see}.
One of these Achilles' heels is when they require cultural context to answer questions~\cite{pawar2025survey}.
This issue is echoed more boldly when analyzing culture with limited internet-based data, such as Iranian culture~\cite{shamsfard2025farseval,hosseinbeigi2025advancing} and Persian language~\cite{rajabi2021survey}.
Benchmarking LLMs on languages and cultures that have been underrepresented in evaluation—such as Persian—is a vital step toward building AI systems capable of engaging more meaningfully and empathetically with diverse user communities.
As LLMs evolve, the development of comprehensive evaluation frameworks, particularly for non-English languages, has become more crucial for robust benchmarking of performance and reliability across diverse linguistic contexts~\cite{hodak2023benchmarking}. 

Our key contributions are as follows:

\noindent\textbf{(I) Curating New Datasets:} We created \textbf{13}  datasets to better evaluate LLMs on Iranian culture and Persian linguistics.

\noindent\textbf{(II) Adapting Well-Known Datasets to Persian:} Beyond translation, we align well-known datasets with Persian language and Iranian cultural context.

\noindent\textbf{(III) Comprehensive Evaluation on Private Test Sets:} We evaluate 41 LLMs to robustly analyze model families and parameter effects, using private test sets to minimize data contamination.

We hope our findings contribute to a deeper understanding of capabilities of LLMs in Persian language and support ongoing efforts to develop better datasets and LLMs in Persian language.

\section{Related work}
% In recent years, the widespread adoption of LLMs across diverse fields has underscored the necessity for their evaluation from multiple perspectives. An early contribution to this effort is the Open LLM Leaderboard by Hugging Face, which benchmarks LLMs in various languages, including Arabic, Chinese, and French.
% Additionally, several datasets, such as MMLU Pro~\cite{wang2024mmluprorobustchallengingmultitask} and ARC~\cite{allenai:arc}, have been created to facilitate evaluation.
% However, as most of these datasets originate in English, there have been significant efforts to translate them to other languages, such as
% Chinese~\citet{he2024cmmu}, European languages~\cite{thellmann2024towards}, and Indian~\cite{kj2025indicmmlu}.

% Since the translation alone is insufficient for comprehensive localization, the focus has shifted to include cultural alignment in the pipeline as well~\cite{zhou2025culture}.
% This approach ensures that datasets are not only linguistically accurate but also culturally relevant within specific contexts.
% Several researchers have embarked on developing culturally aligned datasets for their less-represented cultures to enhance the applicability of LLMs, such as for Arabic~\cite{qian2024cameleval,nacar2025towards}, Korean~\cite{kim2024click}, and Russian~\cite{vasilev2025ruscode} cultures.
The rapid adoption of LLMs across domains has highlighted the need for their evaluation from diverse linguistic and cultural perspectives. Early efforts like Hugging Face’s Open LLM Leaderboard benchmark models in multiple languages~\cite{lai2023open}, and evaluation datasets such as MMLU-pro~\cite{wang2024mmluprorobustchallengingmultitask} and ARC~\cite{allenai:arc} have spurred translation-based adaptations for Chinese~\cite{he2024cmmu}, European~\cite{thellmann2024towards}, and Indian~\cite{kj2025indicmmlu} languages. Yet, translation alone falls short of full localization, prompting a shift toward culturally grounded benchmarks~\cite{zhou2025culture} that ensure both linguistic accuracy and cultural relevance. Recent efforts have introduced culturally aligned datasets for underrepresented communities, including Arabic~\cite{qian2024cameleval,nacar2025towards}, Korean~\cite{kim2024click}, and Russian~\cite{vasilev2025ruscode}, paving the way for more equitable and context-aware LLM evaluation.

Building on the efforts to adapt datasets for different languages and cultures, the Persian language has also seen advancements in the development of resources for LLM training~\cite{hosseinbeigi2025matina,sabouri2022naab} and evaluation~\cite{hosseinbeigi2025advancing,shamsfard2025farseval}.
Parsbench \cite{parsbench2025} has emerged as the first Persian leaderboard, specifically evaluating LLMs using translations of well-known English datasets.
Furthermore, \citet{farsi2025persian} has developed the first benchmark for visual language models in Persian by generating five new datasets.
\citet{ghahroodi2024khayyam} have also contributed by producing a Persian version of the MMLU datasets, encompassing 38 diverse tasks with 20,192 four-choice questions extracted from Persian examinations.
Moreover, \citet{hosseinbeigi2025advancing} has advanced Persian language and cultural benchmarking through the introduction of two new datasets: PeKA, a compilation of Persian mobile application games with diverse, user-generated questions, and PK-BETS, focusing on Persian knowledge and bias ethics categories, albeit with a relatively small sample size of 4,000.

The work most closely related to ours is that of \citet{shamsfard2025farseval}, who also look at the LLM benchmarking problem from the cultural perspective.
They curated a relatively small 4,000 question-answer pairs, including topics like medicine, law, religion, social knowledge, ethics, and bias specific to Iranian culture. The questions were in the form of multiple-choice answers as well as open text generation. They benchmarked three LLMs, including Llama3-70B, and two other Farsi-specific LLMs on their benchmark.
While useful, they didn't utilize well-established current datasets in English.

% Our work fills the gap in two ways: (i) by creating new datasets grounded in Persian linguistics and Iranian culture, and (ii) by utilizing well-known datasets in a two-stage alignment process that combines translation and cultural localization. The resulting benchmark, Evalestan,\Part{correct the name}  assesses LLMs on tasks that demand a nuanced understanding of both the Iranian culture and the Farsi language.

%\input{sections/benchmark}

\section{Benchmark}
As discussed earlier, this research aims to create a benchmark that enhances understanding of LLM capabilities, focusing on not only the Persian language but also Iranian culture, particularly within the Iranian context. Our contributions are divided into two main categories. (i) Creating New Original Datasets. (ii) Translation and Localization of Available Datasets. Detailed information about all datasets and their categories is presented in Table\ref{tab:datasets}.

\begin{table*}[t]
    \centering
    \footnotesize
    % \resizebox{\textwidth}{!}{

    % \setlength{\tabcolsep}{4pt}
    % \renewcommand{\arraystretch}{1.0}
    \setlength{\tabcolsep}{14pt}
    \begin{tabular}{l c c c c}
        \hline
        \textbf{Dataset} & \textbf{Category} & \textbf{Field} & \textbf{Metric} & \textbf{\#Samples} \\
        \hline
        Parsi-Lit & Original & Persian Linguistic & Accuracy & 777 \\
        DC-Homograph & Original & Persian Linguistic & Accuracy & 108 \\
        MC-Homograph & Original & Persian Linguistic & Accuracy & 434 \\
        Proverbs-Quiz & Original & Persian Linguistic & Accuracy & 370 \\
        Verb-Eval & Original & Persian Linguistic & Accuracy & 3,567 \\
        Religion-Rules & Original & Persian Legals & Accuracy & 175 \\
        Iran-Law & Original & Persian Legals & Accuracy & 300 \\
        Persian-Hellaswag & Original & Common Sense Reasoning & Accuracy & 1,361 \\
        Expert-Eval & Original & Domain Specific Knowledge & Accuracy & 49,669 \\
        ReadingCompQA-text & Original & Reading Comprehension QA & F1-Score & 1,000 \\
        ReadingCompQA-y/n & Original & Reading Comprehension QA & Accuracy & 1,000 \\
        Multiple-Wiki & Original & General Knowledge & Accuracy & 1,000 \\
        ParsTrivia & Original & General Knowledge & Accuracy & 392 \\
        MMLU-pro & Translated & General Knowledge & Accuracy & 1,000 \\
        PIQA & Translated & General Knowledge & Accuracy & 999 \\
        Arc-Challenge & Translated & Common Sense Reasoning & Accuracy & 936 \\
        Arc-Easy & Translated & Common Sense Reasoning & Accuracy & 935 \\
        Winogrande & Translated & Common Sense Reasoning & Accuracy & 1,129 \\
        GSM & Translated & Domain Specific Knowledge & Exact-Match & 1,000 \\
        \hline
    \end{tabular}
    % }
    \caption{Overview of datasets that we create in this research.}
    \label{tab:datasets}
\end{table*}
\subsection{Creating New Original Datasets}
A key challenge in evaluating LLMs is the potential overlap between their training and testing data \cite{zhou2023don}. Research suggests that some LLMs may achieve inflated scores due to the public availability of these datasets \cite{singh2025leaderboard}. Therefore, creating new datasets for benchmarking and keeping them private is essential. Furthermore, aspects such as legal regulations, cultural norms, and religious rules are often specific to individual countries and vary significantly from one to another. Consequently, it is crucial to develop new datasets that encompass these unique elements.

In this research, we introduce \textbf{13} new datasets crafted to encompass various aspects, including legal systems, stereotypes, religious inquiries, literature, and more. In what follows, we provide a description of each of these datasets.

\noindent\textbf{Multiple-Wiki:}
This dataset consists of 1,000 multiple-choice questions extracted from the SynTran-Fa dataset \citep{farsi2024syntran} which is a dataset about general knowledge. Questions that did not meet the criteria outlined by \cite{wei2024measuring} were eliminated. Subsequently, incorrect answer options were manually created by one of the co-authors of this paper. The reliability of these questions was then verified by two undergraduate students.

\noindent\textbf{Parsi-Lit:}
Persian language possesses a rich literary heritage encompassing diverse forms of poetry, prose, and classical texts. Building on this cultural wealth, we developed a dataset containing multiple-choice questions sourced from Persian literature curriculum spanning grades 7 through 12. This educational dataset captures the unique linguistic and literary elements characteristic of Persian literary tradition.

\noindent\textbf{Iran-Law:}
To evaluate LLMs understanding of country-specific regulations, we introduce Iran-Law, a dataset comprising multiple-choice questions focused on Iranian legal frameworks. The dataset was developed through a rigorous process involving three domain experts, each holding a PhD in legal studies. Each expert crafted different questions covering diverse aspects of Iranian legislation. To ensure quality and accuracy, we implemented a cross-validation process where experts reviewed each other's questions, establishing a comprehensive evaluation framework for assessing models legal domain knowledge.

\noindent\textbf{Religion-Rules:}
We present a comprehensive dataset addressing religious diversity in Iran, encompassing multiple faiths: Islam (both Shi'a and Sunni), and Zoroastrianism. To ensure authenticity and accuracy in religious content, we collaborated with clergymen from each faith tradition to develop original multiple-choice questions. The dataset comprises various questions distributed as follows: questions covering Islamic jurisprudence (Shi'a and Sunni traditions) and questions for Zoroastrian religious practices. This expert-driven approach was chosen over translation-based methods to maintain doctrinal precision and cultural sensitivity.

\noindent\textbf{Verb-Eval:}
We introduce Verb-Eval, a comprehensive dataset designed to evaluate LLMs on their understanding of Persian verb grammar. This dataset, seeded with an initial collection of approximately 10,000 Persian simple and compound verbs \citep{10.1007/978-3-642-25324-9_34}, served as a foundation for creating the evaluation set. To ensure quality, we filtered out uncommon verbs and selectively sampled compound verbs sharing the same simple root. Using automated scripts, we generated verb forms across various tenses, pronouns, and passive structures, organized into seven distinct linguistic tasks. Two tasks focus on identification: TenseDetection (recognizing a verb's tense) and InfinitiveDetection (finding the correct infinitive). Another task, VerbDetection, assesses conjugation by asking the model to produce a specific verb form from an infinitive based on tense, pronoun, count, and definiteness. Two transformation tasks evaluate morphological manipulation: TenseTransform, which modifies a verb's tense while holding other features constant, and PronounTransform, which modifies the pronoun and count while keeping the tense fixed. The final tasks, TransitiveDetection and VerbTypeDetection, test the models ability to classify a verb's transitivity and its structural type (e.g., simple, compound). This benchmark offers valuable insights into the capabilities of LLMs and their tokenizers in analyzing the structural complexities of Persian verbs.

\noindent\textbf{Proverbs-Quiz:}
Proverbs-Quiz was developed by collecting a seed set of 370 unique and widely used proverbs in Persian literature and everyday language and the meaning of each one from online sources. Each question in the dataset presents a proverb as context, with four answer options randomly selected from the meanings of other proverbs in the seed data. This design enables the assessment of LLMs understanding of Persian idioms and figurative expressions, which are essential for comprehending and generating culturally rich texts.

\noindent\textbf{MC-Homograph:}
Recognizing homographs—words with identical spelling but different meanings—is crucial for clear Persian communication, preventing ambiguity. The Multiple Choice-Homograph dataset is an evaluation set featuring four-option questions. An expert compiled Persian homographs, including their phonemes, meanings, and example contexts, to create this set. Each question presents a homograph within a contextual sentence, requiring users to select its correct meaning from the provided options. This dataset assesses a models ability to accurately interpret homographs in specific contexts.

\noindent\textbf{DC-Homograph:}
The Dual-Context Homograph dataset presents a more complex challenge compared to the Multiple Choice-Homograph dataset. It was developed using the existing collection of Persian homographs, with a LLM prompted to create contexts that incorporate both meanings of each homograph. Prompts included the homograph’s phoneme, meanings, and example usage to ensure the LLM generated accurate and relevant samples. Human reviewers then refined the contexts through multiple rounds of editing or removing unsuitable entries to produce the final evaluation set. The dataset consists of two-option questions, tasking models with identifying the intended meaning of either the first or second instance of the homograph based on subtle contextual cues. This dataset thoroughly evaluates advanced understanding of Persian homographs and underscores the complexities of resolving lexical ambiguity.

\noindent\textbf{ParsTrivia:}
ParsTrivia is a four-choice multiple-choice dataset designed to evaluate the general knowledge capabilities of language models in Persian. The questions in this dataset reflect what are commonly known in Iran as {\small\<سوالات اطلاعات عمومی>} ("general knowledge questions"), which are widely used in quizzes, competitions, and educational contexts. The data was collected by crawling general knowledge questions available on various persian websites. This multi-domain benchmark provides a broad assessment of a models ability to comprehend and respond to diverse factual knowledge questions in Persian.

\noindent\textbf{Expert-Eval:}
Expert-Eval is a specialized benchmark designed to evaluate the expert-level knowledge of language models across three academic tiers: Olympiad-level, Master's-level, and Ph.D.-level questions. To construct the dataset, we collected questions from several reputable Iranian examination sources, including national student Olympiads, the Master's entrance exam, the Ph.D. entrance exam, and two specialized professional assessments: the Legal Apprenticeship License Exam ({\small\<آزمون پذيرش متقاضيان پروانه کارآموزی وکالت>}) and the Professional Competency Exam for Psychologists and Counselors ({\small\<آزمون صلاحيت حرفه‌ای روان‌شناسان و مشاوران>}). All materials were obtained from publicly available PDFs or image files on the internet. Since the original content was not in an editable format, a team of 30 typists manually transcribed the questions. Additionally, all mathematical and scientific formulas were typeset in LaTeX to ensure clarity and to align with the input format best understood by LLMs.
The dataset spans a wide range of domains, including mathematics, engineering, law, psychology, medicine, history, the Persian language, and more. Inspired by the structure of the MMLU dataset \citep{hendryckstest2021}, Expert-Eval is organized into four categories: Humanities, Social Sciences, STEM, and Others. The questions are presented in a multiple-choice format, typically with four options, though some extend to five, allowing for fine-grained evaluation of LLMs advanced capabilities and subject-matter expertise across diverse disciplines.

\noindent\textbf{Persian-Hellaswag:}
Persian-Hellaswag is a multiple-choice benchmark designed to evaluate the ability of language models to predict the most plausible continuation of a given context in Persian. Adapted from the original English HellaSwag dataset, this Persian variant focuses on commonsense reasoning and narrative completion. Since one of the main sources cited in the original paper was the WikiHow website, we also used the Persian version of this site\footnote{\url{https://www.wikihowfarsi.com}} and crawled it accordingly. We then constructed sentence continuation questions based on the articles from this site, as it provides step-by-step explanations for performing various tasks. Each instance presents a short context followed by four candidate continuations, from which the model must select the most coherent and contextually appropriate ending. This benchmark tests models commonsense reasoning and coherence generation in Persian, offering insights into their contextual understanding and narrative prediction capabilities.

\noindent\textbf{ReadingCompQA-text:}
We introduce ReadingCompQA-text, a dataset comprising 1,000 questions designed to evaluate LLMs reading comprehension abilities. Each question is generated from a unique text passage, ensuring broad topical coverage and diversity. Answers are explicitly present within the source text and can be precisely identified using character index spans, facilitating exact answer localization. This design allows for both span extraction and answer generation tasks, providing a clear framework for evaluating models comprehension skills. The dataset serves as a practical benchmark for assessing models ability to process, understand, and retrieve factual information from textual content.

\noindent\textbf{ReadingCompQA-y/n:}
ReadingCompQA-y/n is a dataset of 1,000 yes/no questions, each derived from a distinct text passage, designed to evaluate LLMs reading comprehension abilities in a binary response setting. Each question targets a specific fact or statement directly inferable from the source text, requiring models to answer strictly with “yes” or “no.” The dataset covers diverse topics to challenge models across varying contexts. By focusing on binary classification based on text understanding, ReadingCompQA-y/n provides a focused evaluation framework for assessing models factual comprehension and reasoning accuracy.

\subsection{Translation and Localization of Available Datasets}

To facilitate a meaningful comparison of LLMs across different languages, we developed new datasets by translating and localizing well-known standard datasets used in prominent leaderboards.\footnote{See: \href{https://huggingface.co/spaces/open-llm-leaderboard/open_llm_leaderboard}{Open English Leaderboard}, \href{https://huggingface.co/spaces/le-leadboard/OpenLLMFrenchLeaderboard}{Open French Leader-
board}, \href{https://huggingface.co/spaces/upstage/open-ko-llm-leaderboard}{Open Korean Leaderboard}, \href{https://huggingface.co/spaces/OALL/Open-Arabic-LLM-Leaderboard}{Open Arabic Leaderboard}}

In our approach to convert English benchmark datasets into Persian localized datasets, we employed a multi-step agnatic workflow similar to the methods used by \citet{robinson-etal-2023-chatgpt, gao2024design, wu2024perhaps}.\newline 
Initially, we used the GPT-o4-mini model to identify words within each instance of the original datasets that required conversion to Persian localized terms. After a manual review, we provided GPT-o4-mini with a dictionary of English words and their corresponding Persian localized equivalents according to the context to facilitate translation into Persian.

Finally, to ensure accuracy and cultural alignment, the translations underwent an additional manual review by three Iranian individuals with at least $C1$ proficiency in English. Experts were consulted for instances requiring specialized knowledge. The inter-rater reliability of these reviews was quantified using Cohen's kappa score, which yielded a score of 0.92, indicating perfect agreement. The detailed evaluation methodologies and results, including scores and evaluators' criteria, are documented in Appendix \ref{app:eval-guide}. This translation and localization process resulted in the following benchmark datasets:

\noindent\textbf{Arc-Easy:} A subset of the AI2 Reasoning Challenge dataset containing elementary-level science questions designed to test basic reasoning and scientific knowledge. These questions are characterized by their straightforward nature and require fundamental scientific understanding.

\noindent\textbf{Arc-Challenge:} The more complex counterpart of ARC Easy, featuring advanced scientific reasoning questions that require sophisticated problem-solving skills, multi-step logical inference, and deeper scientific knowledge. These questions are specifically selected for their difficulty in being answered through simple text matching or retrieval.

\noindent\textbf{MMLU-pro:} An advanced version of the Massive Multitask Language Understanding benchmark, covering professional-level knowledge across various fields including law, medicine, engineering, and business. This dataset tests models capabilities in specialized professional domains requiring expert-level understanding.

\noindent\textbf{GSM:} (Grade School Math) A collection of high-quality mathematics word problems that target grade-school math reasoning. These problems require multi-step problem-solving abilities and test models capacity for mathematical reasoning in practical contexts.

\noindent\textbf{PIQA:}
PIQA (\textbf{P}hysical \textbf{I}nteraction \textbf{Q}uestion \textbf{A}nswering) dataset is designed to evaluate a models ability to reason about everyday physical commonsense. The dataset was originally inspired by instructables website.\footnote{\url{https://www.instructables.com}} Each instance consists of a goal and two possible solutions, testing the models understanding of how objects and actions interact in the physical world.

\noindent\textbf{Winogrande:}
Winogrande dataset is a benchmark designed to evaluate commonsense reasoning through pronoun resolution tasks that are challenging for language models. It is a reformulation of the original Winograd Schema Challenge, offering a larger and more diverse set of sentence pairs that require understanding subtle contextual cues to determine the correct referent of a pronoun. Each example presents a sentence with a blank and two candidate options, where only one leads to a coherent and commonsense interpretation.

\section{Evaluation Protocol}
We evaluated all of our introduced datasets using 41 well-known models that have demonstrated good performance in the Persian language. These models range from those fine-tuned specifically on Persian language, such as Maral \citep{https://doi.org/10.57967/hf/5413}, PersianMind \citep{persianmind}, Dorna \citep{part_dp_ai_Dorna}, Dorna2 \citep{part_dp_ai_Dorna2} and Hormoz\footnote{\href{https://huggingface.co/mann-e/Hormoz-8B}{Hormoz LLM}} to well-established multilingual LLMs across a range of parameters, including GPT family \citep{hurst2024gpt, gpt41_web, achiam2023gpt}, Gemini-2 family \citep{gemini_update_2024}, Gemma family \citep{team2025gemma3, team2024gemma2}, Qwen family \citep{yang2025qwen3, Yang2024Qwen25TR, Yang2024Qwen2TR}, LLaMA family \citep{grattafiori2024llama, llama32_2024}, Hermes-3 \citep{teknium2024hermes} and the Cohere family \citep{aryabumi2024aya, dang2024aya}.

\subsection{Evaluation Methodology}

We adopted the evaluation methodology from EleutherAI's \texttt{lm-evaluation-harness} \citep{eval-harness}, extending it to support both local and API-based model settings. We evaluated open-source models hosted locally, as well as proprietary models accessed via APIs. The methodology varies depending on the model type and task category—either multiple-choice questions or open-ended generation.

For locally hosted models, we utilize vLLM \citep{kwon2023efficient} for serving. All bfloat16 and float32 checkpoints are served using bfloat16 precision. Model checkpoints available only in float16 precision (such as some Cohere family models) are served using float16. For multiple-choice tasks, we follow the log-probability-based approach used in \texttt{lm-evaluation-harness}. For each sample, every option is appended to the input prompt, and the models log-probability for the corresponding tokens is computed. The total score for the \( i \)-th option, is calculated as:
\[
\sum_{j=m}^{n_i - 1} \log \mathbb{P}(x_j \mid x_{0:j})
\]
Where \( x_{0:m} \) is the input prompt and \( x_{m:n_i} \) is the \( i \)-th possible option \citep{eleutherai2021multiple}. The option with the highest total log-probability is selected as the models prediction for sample \( k \):
\[
\hat{y}_k = \arg\max_{i \in \{1, 2, \ldots, O_k\}} \sum_{j=m}^{n_i - 1} \log \mathbb{P}(x_j \mid x_{0:j})
\]
Where \( O_k \) is the number of options for sample \( k \). 
The accuracy for sample \( k \) is computed as the indicator function of  $\hat{y}_k = y_k$:
\[
s_k = \mathbf{1}_{ = y_k}(\hat{y}_k) = \begin{cases} 
1 & \text{if } \hat{y}_k = y_k \\
0 & \text{otherwise}
\end{cases}
\]
This process is repeated for all samples, and the overall accuracy is averaged across \( N \) samples:
\[
\text{Accuracy} = \frac{1}{N} \sum_{k=1}^{N} s_k
\]

For generative tasks, we used standard completions and applied robust regex patterns to extract the final answer from the generated output. Performance is then measured using exact match and F1 scores against the reference answers.

For proprietary models accessed via API, we employ a different strategy for choice-based tasks. We leverage structured output features to force the model to generate a JSON object with a "best\_answer" key, where the value is restricted to one of the valid options. The extracted answer is then compared to the target for accuracy scoring. Evaluation of generative tasks for API-based models follows the same procedure as for locally hosted models: we request a standard completion, extract the answer via regex, and evaluate it using exact match or F1.

Because all models we evaluated were instruction-tuned, all multiple-choice tasks are evaluated in a zero-shot setting without any system prompt. For generative tasks, we use a 3-shot context to guide the models toward generating only the final answer, avoiding unnecessary continuation or explanation. For all multiple-choice questions, we enumerate options numerically (e.g., 1, 2, 3, 4). We also examined the effect of using alphabetical option identifiers (e.g., A, B, C, D) instead of numerical ones, and found that this variation has minimal impact on model performance (see Appendix \ref{app:option-format} for details). To ensure reproducibility, all evaluations were conducted with a temperature of 0 to ensure deterministic output.

\subsection{Evaluation Metrics}

Our evaluation metrics are categorized based on the task type. For generation tasks, we used the exact match evaluation method alongside F1 score. For tasks involving multiple-choice questions, accuracy served as the primary metric.

% To further deepen our understanding of the models' confidence levels in multiple-choice and yes/no predictions, we performed an additional analysis that measured uncertainty. This was quantified via normalized Shannon entropy \cite{6773024} applied to the output distribution of the open-source LLMs. The uncertainty score is calculated as shown in Equation \ref{eq:uncer}, where $x_i$ represents the log probability of the choice $i^{th}$ and $n$ is the number of choices for a particular question.

% \begin{equation}
% \centering
% \text{Uncertainty-score} = \frac{-\sum_{i=1}^{n} e^{-x_i} \log e^{-x_i}}{\log n}
% \label{eq:uncer}
% \end{equation}

%\input{sections/results}

\section{Results and Discussion}
%     \begin{figure*}[h]
%             \centering
%     \includegraphics[scale=0.33]{figures/accuracy_distribution.png}
%             \caption{the horizontal bar represents
% the median}
%             \label{fig:results}
%         \end{figure*}

To enhance the interpretability of evaluating LLMs, we categorized the datasets into the following groups:\newline
\textbf{Persian Linguistic:} This category includes datasets such as Verb-Eval, MC-Homograph, DC-Homograph, Proverbs-Quiz and Parsi-Lit.
\newline\textbf{Persian Legals:} This category includes datasets such as Iran-Law, and Religion-Rules.
\newline\textbf{Reading Comprehension QA:} This category includes datasets such as ReadingCompQA-y/n, and ReadingCompQA-text.
\newline\textbf{General Knowledge:} This group comprises datasets including ParsTrivia, PIQA, MMLU-pro, and Multiple-Wiki.
\newline\textbf{Domain Specific Knowledge:} Encompasses datasets like Expert-Eval and GSM.
\newline\textbf{Common Sense Reasoning:} Contains datasets such as Winogrande, Persian-Hellaswag, ARC-Easy, and ARC-Challenge.
\newline
We evaluated 41 LLMs across these categories, as shown in Table \ref{tab:category-average-results}, which presents the performance of these models across different categories. The results are detailed in terms of accuracy for multiple-choice datasets and F1 scores for generation tasks. The findings underscore a significant need for improvement in the models ability to manage the Persian language. Even those models fine-tuned with Persian corpora demonstrate notable performance limitations. Moreover, closed-source models consistently outperform open-source ones, underscoring their effectiveness in managing these datasets. Notably, performance was particularly poor on datasets specifically focused on Iranian culture and the Persian language, with results significantly lower compared to those achieved on other datasets.

To assess LLMs familiarity with Iranian culture, we examined their performance on Persian-Iranian specific datasets. The results show that all LLMs performed poorly, with only one model among the 41 achieving over 50\% accuracy, indicating limited cultural understanding and the need for new, culturally aligned datasets.

%\textbf{Are Models fall in trap or they are not familiar?}
%like \citet{kim2024click} we messure the uncerteinity of model using shanon Enthroupy, and then we define questions that models certeinity is over 50 percent and choose the wring answer, as an trap

% \input{tables/original datasets results}
% \input{tables/translated localized datasets results}

%\input{sections/discussion}

%\input{sections/conclusion}
\section{Conclusion}
In this study, we addressed the existing gap in evaluating large language models (LLMs) for the Persian language by introducing several new datasets.
These datasets fall into two categories: (i) translations and localizations of well-known datasets adapted to Iranian culture and the Persian language, and (ii) datasets newly created specifically for this purpose. Together, these datasets comprehensively cover all aspects of LLM usage in this linguistic context.

Our results reveal that, currently, the OpenAI model family outperforms others on tasks involving the Persian language and demonstrates a comparatively better understanding of Iranian contexts. However, even these models—similar to others, including those specifically fine-tuned for Persian—display performance weaknesses on tasks that are distinctly Iranian or Persian-specific. 

Furthermore, our experiments investigating the impact of model size show a positive correlation between the number of parameters in LLMs and their performance within each model family. %Despite this, there appears to be no consistent pattern in their uncertainty scores. 

\section{Limitations}
Our study has two main limitations regarding the scope of our datasets and evaluation tasks.

First, the creation of our new benchmarks involved inherent challenges. For a large-scale dataset like Expert-Eval, the meticulous process of transcribing thousands of samples from static documents means that despite a careful verification workflow, the potential for occasional minor errors or inconsistencies remains. The scale of this dataset provides a broad measure of model performance, which is intended to complement the specific cultural insights gained from our smaller datasets. Furthermore, our findings are primarily shaped by the Iranian context represented in the data; we believe future work would benefit significantly from expanding this scope to include broader Persian-speaking communities.

Second, the study's focus on multiple-choice and specific generative tasks may not fully capture the complete range of an LLMs capabilities. Important skills such as long-form text generation and dialogue coherence in Persian were not assessed. These areas present valuable avenues for future investigation and benchmark development.
\begin{table*}[t]
    \centering
    % \scriptsize
    \setlength{\aboverulesep}{0pt}
    \setlength{\belowrulesep}{0pt}
    \setlength{\tabcolsep}{24.6pt}
    \resizebox{\linewidth}{!}{
    \begin{tabular}{lccccccc} 
        \toprule
        \textbf{Model} & \textbf{Average} & \textbf{PLing} & \textbf{PLeg} & \textbf{RCQA} & \textbf{GK} & \textbf{DSK} & \textbf{CSR} \\
        \midrule
        \multicolumn{8}{c}{\small\textbf{GPT Models}} \\
        \midrule
        gpt-4o-2024-08-06 & \biggestnumber{72.61} & \biggestnumber{82.25} & 47.27 & 74.72 & 75.18 & \biggestnumber{65.23} & \biggestnumber{91.01} \\
        gpt-4.1-2025-04-14 & 68.98 & 81.84 & \biggestnumber{52.84} & 69.76 & \biggestnumber{75.96} & 42.61 & 90.89 \\
        gpt-4.1-mini-2025-04-14 & 66.71 & 72.61 & 41.02 & 72.78 & 68.33 & 57.34 & 88.17 \\
        gpt-4-turbo-2024-04-09 & 66.14 & 76.93 & 41.00 & \biggestnumber{79.69} & 69.44 & 41.89 & 87.90 \\
        gpt-4o-mini-2024-07-18 & 64.89 & 71.56 & 37.86 & 78.30 & 64.89 & 51.72 & 85.00 \\
        gpt-4.1-nano-2025-04-14 & 57.25 & 59.53 & 32.34 & 66.18 & 57.17 & 50.45 & 77.81 \\
        \midrule
        \multicolumn{8}{c}{\small\textbf{Gemini Models}} \\
        \midrule
        gemini-2.0-flash-001 & 70.68 & 79.13 & 49.41 & 79.61 & 71.76 & 56.60 & 87.57 \\
        gemini-2.0-flash-lite-001 & 65.72 & 72.92 & 42.36 & 79.26 & 67.29 & 46.93 & 85.58 \\
        \midrule
        \multicolumn{8}{c}{\small\textbf{Qwen Models}} \\
        \midrule
        QwQ-32B-Preview & 60.37 & 59.84 & 42.07 & 76.94 & 58.11 & 41.05 & 84.20 \\
        Qwen3-32B & 60.31 & 60.31 & 33.36 & 77.53 & 61.60 & 43.98 & 85.08 \\
        Qwen2.5-32B-Instruct & 59.40 & 63.46 & 40.60 & 60.76 & 58.38 & 48.44 & 84.79 \\
        QwQ-32B & 57.81 & 59.16 & 39.86 & 69.38 & 57.67 & 38.03 & 82.76 \\
        Qwen3-30B-A3B & 55.70 & 55.85 & 29.38 & 76.22 & 55.46 & 34.97 & 82.34 \\
        Qwen3-14B & 54.46 & 57.33 & 31.91 & 65.98 & 53.47 & 37.16 & 80.90 \\
        Qwen2-57B-A14B-Instruct & 53.14 & 57.15 & 29.33 & 71.47 & 53.26 & 31.15 & 76.45 \\
        Qwen3-8B & 53.11 & 53.38 & 28.84 & 76.39 & 50.58 & 32.01 & 77.48 \\
        Qwen2.5-14B-Instruct & 53.09 & 57.16 & 33.00 & 56.78 & 55.44 & 41.11 & 75.09 \\
        Qwen2.5-7B-Instruct & 50.69 & 51.12 & 34.45 & 70.47 & 47.85 & 27.62 & 72.64 \\
        Qwen3-4B & 49.28 & 46.72 & 30.31 & 72.67 & 44.86 & 28.69 & 72.41 \\
        Qwen2-7B-Instruct & 47.80 & 50.00 & 28.17 & 64.57 & 46.80 & 25.41 & 71.88 \\
        Qwen2.5-3B-Instruct & 42.92 & 47.01 & 27.45 & 52.66 & 40.68 & 33.73 & 55.98 \\
        \midrule
        \multicolumn{8}{c}{\small\textbf{Gemma Models}} \\
        \midrule
        gemma-3-27b-it & 60.27 & 68.43 & 30.45 & 74.71 & 63.20 & 38.81 & 86.02 \\
        gemma-2-27b-it & 58.39 & 64.40 & 29.34 & 73.28 & 61.38 & 36.65 & 85.28 \\
        gemma-2-9b-it & 56.47 & 63.07 & 31.69 & 73.07 & 58.21 & 30.22 & 82.57 \\
        gemma-3-12b-it & 57.35 & 67.04 & 30.74 & 71.43 & 59.29 & 32.16 & 83.43 \\
        gemma-3-4b-it & 45.40 & 48.90 & 24.41 & 62.94 & 46.00 & 22.15 & 67.99 \\
        gemma-2-2b-it & 42.34 & 46.90 & 29.19 & 57.10 & 38.72 & 18.86 & 63.25 \\
        gemma-3-1b-it & 34.18 & 36.36 & 24.17 & 47.94 & 31.62 & 15.76 & 49.26 \\
        \midrule
        \multicolumn{8}{c}{\small\textbf{Cohere Models}} \\
        \midrule
        aya-expanse-32b & 59.62 & 64.85 & 37.91 & 78.48 & 62.93 & 30.90 & 82.68 \\
        aya-23-35B & 53.53 & 57.09 & 31.15 & 74.51 & 56.39 & 23.72 & 78.35 \\
        aya-expanse-8b & 51.11 & 55.01 & 31.02 & 72.14 & 51.64 & 22.68 & 74.15 \\
        aya-23-8B & 47.00 & 48.79 & 28.74 & 66.31 & 48.97 & 19.72 & 69.51 \\
        \midrule
        \multicolumn{8}{c}{\small\textbf{Persian Models}} \\
        \midrule
        Hormoz-8B & 50.49 & 54.37 & 29.79 & 70.41 & 51.76 & 22.80 & 73.84 \\
        Dorna2-Llama3.1-8B-Instruct & 47.70 & 45.91 & 31.69 & 69.32 & 45.60 & 23.78 & 69.93 \\
        Dorna-Llama3-8B-Instruct & 45.32 & 42.62 & 27.24 & 72.48 & 41.60 & 22.40 & 65.62 \\
        PersianMind-v1.0 & 35.08 & 39.19 & 26.81 & 33.15 & 35.24 & 16.03 & 60.07 \\
        Maral-7B-alpha-1 & 34.71 & 33.78 & 26.31 & 52.37 & 31.70 & 16.60 & 47.50 \\
        \midrule
        \multicolumn{8}{c}{\small\textbf{Hermes Model \& Llama Models}} \\
        \midrule
        Llama-3.1-8B-Instruct & 49.45 & 49.07 & 31.19 & 72.58 & 48.31 & 24.81 & 70.76 \\
        Meta-Llama-3-8B-Instruct & 48.25 & 46.93 & 32.86 & 68.65 & 48.50 & 23.35 & 69.23 \\
        Hermes-3-Llama-3.1-8B & 48.13 & 50.31 & 30.98 & 69.95 & 46.52 & 22.91 & 68.13 \\
        Llama-3.2-1B-Instruct & 35.30 & 37.05 & 27.72 & 51.05 & 32.31 & 16.35 & 47.34 \\
        \bottomrule
    \end{tabular}
    }
    \caption{Performance results of LLMs on different dataset categories. The model families are ranked by their top-performing model, and within each family, models are sorted by their average performance. The best performance in each column is shown in bold. Abbreviations used: PLing (Persian Linguistic), PLeg (Persian Legals), RCQA (Reading Comprehension QA), GK (General Knowledge), DSK (Domain Specific Knowledge), CSR (Common Sense Reasoning).}
    \label{tab:category-average-results}
\end{table*}
\bibliography{custom}

\clearpage

\appendix

\section{Sampling Method}
For selecting instances from datasets, particularly those with sub-categories like MMLU-pro and ARC, we adhere to the original dataset proportions. To ensure diversity, we utilize k-means clustering on the dataset instances based on their embeddings generated using BERT. The optimal number of clusters, $k$, is determined via the elbow method. Samples are then drawn from each cluster proportional to its size, enhancing the representativeness of our selection.

\section{Complete Results} 
Tables \ref{tab:translated-localized} and \ref{tab:original-data-res} present the performance of 41 models on translated/localized datasets and original, respectively.

\section{Option Format Comparison}
\label{app:option-format}
To investigate whether language models exhibit any bias toward specific option formats, we conducted an experiment comparing numerical and alphabetical option identifiers. Specifically, we selected one small and one large model from each of three model families—Cohere, Qwen, and Gemma—to examine potential biases across both small- and large-scale models within each family. The six models evaluated were \textbf{aya-expanse-32b}, \textbf{aya-expanse-8b}, \textbf{gemma-3-27b-it}, \textbf{gemma-3-12b-it}, \textbf{Qwen3-32B} and \textbf{Qwen3-8B}.

We evaluated the models on three multiple-choice benchmarks: \textbf{ARC-Challenge}, \textbf{ARC-Easy} and \textbf{Winogrande}. These tasks were selected because their English leaderboard versions typically use alphabetical option formats (e.g., ‘A’, ‘B’, ‘C’, etc.), making them suitable for assessing the effect of switching from numerical (1–10) to alphabetical (A–J) identifiers.

As shown in Table  \ref{tab:option}, the average accuracy differences between the two formats are minimal across all models. Most models exhibit differences of around 1\%, with the exception of Qwen3-32B, which shows a slightly larger variation of approximately 2\%. Overall, these results suggest that model performance is generally consistent regardless of whether numerical or alphabetical option formats are used, indicating no strong format bias.

\section{Evaluation Guideline}
\label{app:eval-guide}
\subsection{Human Evaluation}
In human evaluation, four evaluators, each a native Persian speaker with a Master's degree and $C1$ proficiency in English, assess translation quality by assigning a score from 0 to 10, where 0 represents the lowest quality and 10 signifies the best quality. The evaluations are based on the following criteria:

\begin{itemize}
    \item \textbf{Ambiguity:} Assign a score of 0 if the translation is ambiguous. 
    \item \textbf{Incorrect Word Translation:} Deduct 2 points for each incorrectly translated word that does not change the sentence’s overall meaning. Assign a score of 0 if the incorrect translation alters the sentence’s meaning. Note that words altered during the localization process are exempt from these deductions.
    \item \textbf{Grammatical Errors:} Deduct 0.5 points for each grammatical error that does not impact the meaning (e.g., incorrect article). 
    \item \textbf{Accuracy:} If the translation preserves the original meaning and clarity, start from a perfect score and adjust according to these guidelines. 
\end{itemize}

\subsection{Automatic Evaluation:}
In line with \citet{bacciu2024dantellm}, we configured the GPT-4 model with a temperature setting of 0 to ensure precise and consistent responses. The evaluation process utilized the same guidelines as those used by our human annotators.

Table \ref{tab:trans} presents the scores assigned by annotators to the translation of datasets.
\begin{table}[h]
    \centering 
    \def\arraystretch{1.25}
    \setlength{\tabcolsep}{12pt} 
    \resizebox{\linewidth}{!}{
    \begin{tabular}{l|c|c} 
    \hline
    \textbf{Dataset name} & \textbf{GPT-4 Score} &  \textbf{Human Score}\\ \hline 
    MMLU-pro & $9.15 \pm 2.16$   &  $9.47 \pm 0.21$
    \\ \hline 
    GSM & $9.64 \pm 2.02$ &  $9.85 \pm 0.28$
    \\ \hline 
    Arc-Easy & $9.73 \pm 0.86$ &  $9.82 \pm 0.13$
    \\ \hline 
    Arc-Challenge & $9.21 \pm 1.13$ &   $9.36 \pm 0.24$
    \\ \hline 
    % Winogarnde &$9.87 \pm 0.94$ &  $9.83 \pm 0.17$
    % \\ \hline 
     \textbf{AVG}    & $9.36 \pm 1.51$ &    $9.60 \pm 0.34$
     \\ \hline 
    \end{tabular}
    }
    \caption{Translation quality results with Human and Automatic Evaluation.}
    \label{tab:trans}
\end{table}

\subsubsection{Quality Assessment of Localization}
To evaluate the quality of localization, we use the following human evaluation criteria:\newline
Assign a score as follows: A score of 0 if there is a word that could be replaced with a Persian equivalent and has not been replaced; a score of 5 if an English word could be replaced with a Persian equivalent but is not replaced with an appropriate word; and a score of 10 if the replacement is done correctly and appropriately. Table \ref{tab:local} presents the scores assigned by annotators to the localization of datasets.
\begin{table}[h]
    \centering
    \def\arraystretch{1.05}
    \setlength{\tabcolsep}{9pt}
    \resizebox{\linewidth}{!}{
    \begin{tabular}{l|p{0.7\linewidth}}
    \hline
    \textbf{Dataset name} &  \textbf{Human Score}\\ \hline
     {MMLU-pro}       & $9.1 \pm 0.09$ \\
     {GSM}            & $9.21 \pm 0.13$ \\
     {Arc-Easy}       & $8.9 \pm 0.11$ \\
     {Arc-Challenge}  & $8.7 \pm 0.16$ \\
     % \textbf{Winogarnde}     & $8.0 \pm 0.17$ \\ \hline
    \textbf{AVG}             & $8.8 \pm 0.14$ \\ \hline
    \end{tabular}
    }
    \caption{Localization quality results with Human Evaluation.}
    \label{tab:local}
\end{table}

\bigskip

\begin{center}
\begin{minipage}{\textwidth}
    \centering
    \resizebox{\linewidth}{!}{
        \begin{tabular}{|l|c|c|c|c|c|c|c|c|}
            \hline
            Model & \multicolumn{2}{c|}{Average} & \multicolumn{2}{c|}{Arc-Challenge} & \multicolumn{2}{c|}{Arc-Easy} & \multicolumn{2}{c|}{Winogrande} \\
            \cline{2-9}
             & Numerical & Alphabetical & Numerical & Alphabetical & Numerical & Alphabetical & Numerical & Alphabetical\\
            \hline
            aya-expanse-32b & 83.01& 83.44& 85.15& 85.04& 93.37& 93.80& 70.50& 71.48\\
            aya-expanse-8b & 73.37& 74.23& 71.47& 73.50& 84.60& 84.71& 64.04& 64.48\\
            gemma-3-27b-it & 86.90& 86.73& 88.35& 89.21& 94.22& 94.55& 78.12& 76.44\\
            gemma-3-12b-it & 83.51& 83.89& 83.33& 84.72& 93.26& 93.69& 73.95& 73.25\\
            Qwen3-32B & 85.61& 83.57& 91.13& 87.50& 94.22& 91.55& 71.48& 71.66\\
            Qwen3-8B & 76.51& 77.70& 80.24& 81.52& 87.38& 87.81& 61.91& 63.77\\
            \hline
        \end{tabular}
    }
    \captionof{table}{Model accuracies under two option formats: Alphabetical (A–J) and Numerical (1–10).}
    \label{tab:option}
\end{minipage}
\end{center}

\begin{table*}[t]
    \centering
    % \footnotesize
    % \setlength{\aboverulesep}{0.3pt}
    % \setlength{\belowrulesep}{0.3pt}
    \resizebox{\linewidth}{!}{
    \begin{tabular}{lccccccc}
        \toprule
        \textbf{Model} & \textbf{Average} & \textbf{Arc-Challenge} & \textbf{Arc-Easy} & \textbf{MMLU-pro} & \textbf{PIQA} & \textbf{GSM} & \textbf{Winogrande} \\
        \midrule
        \multicolumn{8}{c}{\textbf{GPT Models}} \\
        \midrule
        gpt-4o-2024-08-06 & \biggestnumber{82.30} & 95.09 & \biggestnumber{97.22} & 47.10 & 95.10 & \biggestnumber{73.10} & \biggestnumber{86.18} \\
        gpt-4.1-mini-2025-04-14 & 78.15 & 91.88 & 96.15 & 47.80 & 92.69 & 60.30 & 80.07 \\
        gpt-4.1-2025-04-14 & 74.93 & \biggestnumber{95.30} & 96.68 & \biggestnumber{50.50} & \biggestnumber{95.90} & 25.30 & 85.92 \\
        gpt-4o-mini-2024-07-18 & 73.79 & 86.43 & 94.01 & 34.80 & 90.89 & 60.90 & 75.73 \\
        gpt-4-turbo-2024-04-09 & 72.63 & 91.35 & 96.47 & 40.10 & 94.19 & 30.60 & 83.08 \\
        gpt-4.1-nano-2025-04-14 & 67.69 & 81.41 & 91.55 & 29.90 & 84.58 & 58.40 & 60.32 \\
        \midrule
        \multicolumn{8}{c}{\textbf{Gemini Models}} \\
        \midrule
        gemini-2.0-flash-001 & 76.57 & 91.35 & \biggestnumber{97.22} & 47.80 & 90.59 & 53.70 & 78.74 \\
        gemini-2.0-flash-lite-001 & 70.83 & 89.64 & 93.48 & 41.20 & 85.29 & 39.70 & 75.64 \\
        \midrule
        \multicolumn{8}{c}{\textbf{Qwen Models}} \\
        \midrule
        Qwen2.5-32B-Instruct & 71.43 & 85.15 & 91.87 & 37.40 & 83.98 & 50.10 & 80.07 \\
        Qwen3-32B & 70.87 & 91.13 & 94.22 & 42.80 & 87.69 & 37.90 & 71.48 \\
        QwQ-32B-Preview & 67.66 & 85.58 & 91.44 & 37.30 & 81.28 & 34.70 & 75.64 \\
        QwQ-32B & 66.47 & 84.94 & 90.80 & 39.00 & 81.68 & 29.30 & 73.07 \\
        Qwen3-14B & 64.40 & 84.29 & 91.02 & 35.50 & 77.18 & 31.10 & 67.32 \\
        Qwen3-30B-A3B & 63.97 & 87.39 & 93.58 & 36.30 & 72.47 & 28.80 & 65.28 \\
        Qwen2.5-14B-Instruct & 61.96 & 82.26 & 87.91 & 34.60 & 76.98 & 38.80 & 51.21 \\
        Qwen2-57B-A14B-Instruct & 58.94 & 76.71 & 85.35 & 27.00 & 76.88 & 22.10 & 65.63 \\
        Qwen2.5-7B-Instruct & 55.25 & 72.33 & 81.50 & 26.70 & 71.07 & 18.00 & 61.91 \\
        Qwen3-4B & 54.44 & 73.61 & 83.42 & 28.90 & 66.07 & 20.10 & 54.56 \\
        Qwen2-7B-Instruct & 53.35 & 69.12 & 80.75 & 23.80 & 70.97 & 14.50 & 60.94 \\
        Qwen2.5-3B-Instruct & 45.80 & 51.50 & 67.27 & 21.10 & 62.16 & 34.10 & 38.66 \\
        \midrule
        \multicolumn{8}{c}{\textbf{Gemma Models}} \\
        \midrule
        gemma-3-27b-it & 68.81 & 88.35 & 94.22 & 36.60 & 87.29 & 28.30 & 78.12 \\
        gemma-2-27b-it & 68.45 & 86.75 & 94.22 & 36.90 & 89.69 & 26.70 & 76.44 \\
        gemma-3-12b-it & 65.09 & 83.33 & 93.26 & 32.60 & 87.19 & 20.20 & 73.96 \\
        gemma-2-9b-it & 64.53 & 84.29 & 93.16 & 33.20 & 87.09 & 17.40 & 72.01 \\
        gemma-3-4b-it & 50.55 & 63.46 & 79.57 & 22.80 & 72.77 & 9.60 & 55.09 \\
        gemma-2-2b-it & 45.77 & 57.91 & 70.48 & 18.20 & 66.87 & 6.40 & 54.74 \\
        gemma-3-1b-it & 34.80 & 36.43 & 46.10 & 13.70 & 57.66 & 4.30 & 50.58 \\
        \midrule
        \multicolumn{8}{c}{\textbf{Cohere Models}} \\
        \midrule
        aya-expanse-32b & 64.97 & 85.15 & 93.37 & 32.10 & 91.19 & 17.50 & 70.50 \\
        aya-23-35B & 59.52 & 77.56 & 90.16 & 24.10 & 89.49 & 10.00 & 65.81 \\
        aya-expanse-8b & 55.33 & 71.47 & 84.60 & 21.90 & 80.18 & 9.80 & 64.04 \\
        aya-23-8B & 51.50 & 63.68 & 81.39 & 19.90 & 80.78 & 6.10 & 57.13 \\
        \midrule
        \multicolumn{8}{c}{\textbf{Persian Models}} \\
        \midrule
        Hormoz-8B & 55.23 & 70.73 & 84.39 & 21.70 & 80.48 & 9.90 & 64.22 \\
        Dorna2-Llama3.1-8B-Instruct & 50.90 & 67.63 & 78.72 & 22.70 & 69.97 & 11.90 & 54.47 \\
        Dorna-Llama3-8B-Instruct & 47.55 & 59.94 & 70.70 & 22.00 & 66.17 & 10.30 & 56.16 \\
        PersianMind-v1.0 & 42.18 & 54.59 & 69.73 & 14.50 & 59.76 & 2.30 & 52.17 \\
        Maral-7B-alpha-1 & 33.78 & 37.29 & 43.10 & 14.80 & 51.95 & 6.10 & 49.42 \\
        \midrule
        \multicolumn{8}{c}{\textbf{Hermes Model \& Llama Models}} \\
        \midrule
        Llama-3.1-8B-Instruct & 51.83 & 68.91 & 80.11 & 25.70 & 70.07 & 12.00 & 54.21 \\
        Meta-Llama-3-8B-Instruct & 51.26 & 66.77 & 76.47 & 26.00 & 70.97 & 10.40 & 56.95 \\
        Hermes-3-Llama-3.1-8B & 50.53 & 65.28 & 78.07 & 24.10 & 70.37 & 10.20 & 55.18 \\
        Llama-3.2-1B-Instruct & 34.63 & 37.50 & 47.38 & 15.70 & 54.05 & 4.10 & 49.07 \\
        \bottomrule
    \end{tabular}
    }
    \caption{Results of LLMs on translated/localized datasets. The model families are ranked by their top-performing model, and within each family, models are sorted by their average performance. The best performance in each column is shown in bold.}
    \label{tab:translated-localized}
\end{table*}

\begin{table*}[t]
    \centering
    \resizebox{\linewidth}{!}{
    \begin{tabular}{lcccccccccccccc}
        \toprule
        \textbf{Model} & \textbf{Average} & \textbf{MW} & \textbf{PL} & \textbf{IL} & \textbf{RR} & \textbf{VE} & \textbf{PQ} & \textbf{MCH} & \textbf{DCH} & \textbf{PT} & \textbf{EE} & \textbf{PH} & \textbf{RC-text} & \textbf{RC-y/n} \\
        \midrule
        \multicolumn{15}{c}{\textbf{GPT Models}} \\
        \midrule
        gpt-4.1-2025-04-14 & \biggestnumber{73.65} & 66.60 & 45.82 & \biggestnumber{53.67} & 52.00 & 83.04 & 95.14 & 95.39 & \biggestnumber{89.81} & \biggestnumber{90.82} & \biggestnumber{59.92} & \biggestnumber{85.67} & 44.82 & \biggestnumber{94.70} \\
        gpt-4o-2024-08-06 & 73.59 & \biggestnumber{67.70} & \biggestnumber{45.95} & 47.67 & 46.86 & \biggestnumber{85.89} & \biggestnumber{96.76} & \biggestnumber{95.62} & 87.04 & \biggestnumber{90.82} & 57.36 & 85.53 & 55.34 & 94.10 \\
        gpt-4-turbo-2024-04-09 & 69.49 & 62.60 & 40.93 & 42.00 & 40.00 & 74.29 & 86.76 & 93.78 & 88.89 & 80.87 & 53.18 & 80.68 & 67.17 & 92.20 \\
        gpt-4.1-mini-2025-04-14 & 66.34 & 53.50 & 41.18 & 44.33 & 37.71 & 77.99 & 82.97 & 94.24 & 66.67 & 79.34 & 54.37 & 84.57 & 51.85 & 93.70 \\
        gpt-4o-mini-2024-07-18 & 65.41 & 54.80 & 40.93 & 34.00 & 41.71 & 74.23 & 84.05 & 90.09 & 68.52 & 79.08 & 42.54 & 83.84 & 63.29 & 93.30 \\
        gpt-4.1-nano-2025-04-14 & 56.10 & 46.10 & 36.42 & 32.67 & 32.00 & 66.21 & 67.84 & 78.11 & 49.07 & 68.11 & 42.49 & 77.96 & 50.66 & 81.70 \\
        \midrule
        \multicolumn{15}{c}{\textbf{Gemini Models}} \\
        \midrule
        gemini-2.0-flash-001 & 72.68 & 60.90 & 44.02 & 45.67 & \biggestnumber{53.14} & 85.15 & 95.14 & 91.71 & 79.63 & 87.76 & 59.50 & 82.95 & \biggestnumber{67.92} & 91.30 \\
        gemini-2.0-flash-lite-001 & 68.32 & 58.50 & 43.89 & 43.00 & 41.71 & 81.39 & 91.35 & 87.79 & 60.19 & 84.18 & 54.15 & 83.54 & 65.92 & 92.60 \\
        \midrule
        \multicolumn{15}{c}{\textbf{Gemma Models}} \\
        \midrule
        gemma-3-27b-it & 62.62 & 55.20 & 40.93 & 36.33 & 24.57 & 66.02 & 78.92 & 92.40 & 63.89 & 73.72 & 49.32 & 83.39 & 58.01 & 91.40 \\
        gemma-3-12b-it & 60.32 & 49.00 & 40.03 & 36.33 & 25.14 & 63.39 & 72.97 & 91.24 & 67.59 & 68.37 & 44.12 & 83.17 & 55.26 & 87.60 \\
        gemma-2-27b-it & 59.73 & 50.80 & 35.91 & 34.67 & 24.00 & 61.16 & 73.51 & 91.24 & 60.19 & 68.11 & 46.60 & 83.69 & 56.76 & 89.80 \\
        gemma-2-9b-it & 58.56 & 48.50 & 38.10 & 33.67 & 29.71 & 58.25 & 69.19 & 90.55 & 59.26 & 64.03 & 43.03 & 80.82 & 56.43 & 89.70 \\
        gemma-3-4b-it & 47.40 & 42.50 & 30.24 & 27.67 & 21.14 & 45.30 & 53.78 & 72.58 & 42.59 & 45.92 & 34.70 & 73.84 & 47.28 & 78.60 \\
        gemma-2-2b-it & 44.47 & 36.90 & 30.76 & 32.67 & 25.71 & 36.18 & 45.68 & 74.65 & 47.22 & 32.91 & 31.31 & 69.88 & 41.79 & 72.40 \\
        gemma-3-1b-it & 36.33 & 29.10 & 24.97 & 20.33 & 28.00 & 27.67 & 28.92 & 51.15 & 49.07 & 26.02 & 27.22 & 63.92 & 31.98 & 63.90 \\
        \midrule
        \multicolumn{15}{c}{\textbf{Cohere Models}} \\
        \midrule
        aya-expanse-32b & 62.42 & 54.70 & 34.75 & 38.67 & 37.14 & 61.95 & 77.03 & 87.56 & 62.96 & 73.72 & 44.29 & 81.70 & 67.25 & 89.70 \\
        aya-23-35B & 55.85 & 48.70 & 31.92 & 32.00 & 30.29 & 47.32 & 67.03 & 83.64 & 55.56 & 63.27 & 37.44 & 79.87 & 62.82 & 86.20 \\
        aya-expanse-8b & 53.68 & 45.80 & 34.49 & 32.33 & 29.71 & 48.06 & 60.00 & 80.65 & 51.85 & 58.67 & 35.56 & 76.49 & 61.98 & 82.30 \\
        aya-23-8B & 49.11 & 42.90 & 31.27 & 28.33 & 29.14 & 39.30 & 44.32 & 76.27 & 52.78 & 52.30 & 33.33 & 75.83 & 60.31 & 72.30 \\
        \midrule
        \multicolumn{15}{c}{\textbf{Qwen Models}} \\
        \midrule
        QwQ-32B-Preview & 60.20 & 50.60 & 39.77 & 43.00 & 41.14 & 51.97 & 58.11 & 88.25 & 61.11 & 63.27 & 47.39 & 84.13 & 65.38 & 88.50 \\
        Qwen3-32B & 59.44 & 48.30 & 39.12 & 37.00 & 29.71 & 56.35 & 64.59 & 89.63 & 51.85 & 67.60 & 50.06 & 83.47 & 63.96 & 91.10 \\
        Qwen2.5-32B-Instruct & 58.54 & 50.40 & 40.41 & 42.33 & 38.86 & 54.58 & 63.24 & 91.47 & 67.59 & 61.73 & 46.78 & 82.07 & 28.11 & 93.40 \\
        QwQ-32B & 57.94 & 49.30 & 37.71 & 38.00 & 41.71 & 52.31 & 59.19 & 88.25 & 58.33 & 60.71 & 46.75 & 82.22 & 50.25 & 88.50 \\
        Qwen3-30B-A3B & 55.98 & 48.00 & 36.55 & 35.33 & 23.43 & 48.09 & 50.81 & 86.41 & 57.41 & 65.05 & 41.13 & 83.10 & 66.24 & 86.20 \\
        Qwen2-57B-A14B-Instruct & 54.99 & 48.20 & 33.85 & 30.67 & 28.00 & 52.31 & 58.11 & 85.02 & 56.48 & 60.97 & 40.20 & 78.10 & 57.74 & 85.20 \\
        Qwen3-14B & 54.45 & 44.80 & 35.39 & 34.67 & 29.14 & 54.36 & 53.78 & 87.56 & 55.56 & 56.38 & 43.22 & 80.97 & 44.36 & 87.60 \\
        Qwen2.5-14B-Instruct & 53.68 & 49.70 & 33.85 & 34.00 & 32.00 & 51.21 & 54.59 & 86.87 & 59.26 & 60.46 & 43.41 & 78.99 & 22.46 & 91.10 \\
        Qwen3-8B & 53.17 & 46.00 & 33.20 & 29.67 & 28.00 & 47.93 & 51.89 & 82.95 & 50.93 & 49.23 & 38.31 & 80.38 & 66.38 & 86.40 \\
        Qwen2.5-7B-Instruct & 51.62 & 42.60 & 31.27 & 32.33 & 36.57 & 44.44 & 47.84 & 79.26 & 52.78 & 51.02 & 37.24 & 74.80 & 58.43 & 82.50 \\
        Qwen2-7B-Instruct & 49.30 & 40.40 & 31.40 & 28.33 & 28.00 & 40.62 & 50.54 & 72.81 & 54.63 & 52.04 & 36.31 & 76.71 & 50.14 & 79.00 \\
        Qwen3-4B & 49.18 & 40.60 & 31.79 & 30.33 & 30.29 & 41.23 & 45.41 & 76.27 & 38.89 & 43.88 & 37.28 & 78.03 & 63.43 & 81.90 \\
        Qwen2.5-3B-Instruct & 44.20 & 38.90 & 29.73 & 30.33 & 24.57 & 38.66 & 41.08 & 70.97 & 54.63 & 40.56 & 33.35 & 66.50 & 40.41 & 64.90 \\
        \midrule
        \multicolumn{15}{c}{\textbf{Persian Models}} \\
        \midrule
        Hormoz-8B & 52.99 & 46.70 & 33.08 & 31.00 & 28.57 & 47.38 & 60.27 & 80.18 & 50.93 & 58.16 & 35.70 & 76.05 & 61.11 & 79.70 \\
        Dorna2-Llama3.1-8B-Instruct & 48.91 & 41.00 & 27.28 & 29.67 & 33.71 & 42.06 & 42.97 & 72.81 & 44.44 & 48.72 & 35.65 & 78.91 & 56.84 & 81.80 \\
        Dorna-Llama3-8B-Instruct & 46.22 & 36.90 & 27.54 & 25.33 & 29.14 & 34.74 & 35.41 & 74.65 & 40.74 & 41.33 & 34.49 & 75.68 & 64.85 & 80.10 \\
        PersianMind-v1.0 & 36.62 & 36.10 & 27.80 & 27.33 & 26.29 & 26.26 & 34.32 & 65.90 & 41.67 & 30.61 & 29.75 & 63.78 & 0.00 & 66.30 \\
        Maral-7B-alpha-1 & 36.43 & 28.40 & 26.77 & 26.33 & 26.29 & 28.96 & 22.16 & 47.47 & 43.52 & 31.63 & 27.10 & 60.18 & 42.04 & 62.70 \\
        \midrule
        \multicolumn{15}{c}{\textbf{Hermes Model \& Llama Models}} \\
        \midrule
        Llama-3.1-8B-Instruct & 51.36 & 44.90 & 32.30 & 32.67 & 29.71 & 42.91 & 47.57 & 79.03 & 43.52 & 52.55 & 37.62 & 79.79 & 62.45 & 82.70 \\
        Hermes-3-Llama-3.1-8B & 50.36 & 42.10 & 30.63 & 31.67 & 30.29 & 48.94 & 47.84 & 79.72 & 44.44 & 49.49 & 35.61 & 73.99 & 56.40 & 83.50 \\
        Meta-Llama-3-8B-Instruct & 49.82 & 45.00 & 29.99 & 32.00 & 33.71 & 38.93 & 42.97 & 81.11 & 41.67 & 52.04 & 36.30 & 76.71 & 54.79 & 82.50 \\
        Llama-3.2-1B-Instruct & 37.40 & 29.90 & 27.03 & 24.00 & 31.43 & 26.11 & 28.65 & 52.53 & 50.93 & 29.59 & 28.59 & 55.40 & 38.00 & 64.10 \\
        \bottomrule
    \end{tabular}
    }
    \caption{Results of LLMs on original datasets. The model families are ranked by their top-performing model, and within each family, models are sorted by their average performance. The best performance in each column is shown in bold. Abbreviations used: MW (Multiple-Wiki), PL (Parsi-Lit), IL (Iran-Law), RR (Religion-Rules), VE (Verb-Eval), PQ (Proverbs-Quiz), MCH (MC-Homograph), DCH (DC-Homograph), PT (ParsTrivia), EE (Expert-Eval), PH (Persian-Hellaswag), RC-text (ReadingCompQA-text), RC-y/n (ReadingCompQA-y/n).}
    \label{tab:original-data-res}
\end{table*}
% \clearpage 
% \vspace{5em}
% \input{figures/PGSM}
% \vspace{5em}
% \input{figures/law questions}

% \clearpage
% \vspace{5em}
% \input{figures/Persian literature}
% \vspace{5em}
% \input{figures/FiqhQuestion}

% \clearpage
% \vspace{5em}
% \input{figures/persain proverbs}
% \vspace{5em}
% \input{figures/arc-challange}

% \clearpage
% \vspace{5em}
% \input{figures/arc-easy}
% \vspace{5em}
% \input{figures/aut-multiple-choice}

% \clearpage
% \vspace{5em}
% \input{figures/part multiple choice}
% \vspace{5em}
% \input{figures/winogrande}

% \clearpage
% \vspace{5em}
% \input{figures/homographs hard}
% \vspace{5em}
% \input{figures/homographs easy}

% \clearpage
% \vspace{5em}
% \input{figures/verbdetection}
% \vspace{5em}
% \input{figures/tense detection}

% \clearpage
% \vspace{5em}
% \input{figures/tense transform}
% \vspace{5em}
% \input{figures/pronoun tranform}

% \clearpage
% \vspace{5em}
% \input{figures/infinitivedetection}
% \vspace{5em}
% \input{figures/verbtypedetection}

% \clearpage
% \input{figures/transitivedetection}
% \vspace{5em}
% \input{figures/mmlu-pro}
% \vspace{5em}

% \clearpage
% \vspace{5em}
% \input{figures/QAYN}
% \clearpage
% \vspace{5em}
% \input{figures/piqa}

% \clearpage
% \input{tables/uncertainty original}
% \clearpage
% \input{tables/uncertainty-trans-loc}
% \clearpage
% \input{tables/uncertaintyWOhallu}
\end{document}